\documentclass[10pt, a4paper]{article}
\usepackage{lrec}
\usepackage{multibib}
\newcites{languageresource}{Language Resources}
\usepackage{graphicx}
\usepackage{subcaption}
\usepackage{tabularx}
\usepackage{soul}
\usepackage{amsmath}
\usepackage{CJKutf8}
\usepackage{mathrsfs}
\usepackage{makecell}
\usepackage{adjustbox}
\usepackage{multirow}
\usepackage{algorithm}
\usepackage{verbatim}
\usepackage{setspace}
\usepackage{float}
\usepackage{algpseudocode}

\usepackage{titlesec}
\titleformat{\section}{\normalfont\large\bf\center}{\thesection.}{1em}{}
\titleformat{\subsection}{\normalfont\SmallTitleFont\bf\raggedright}{\thesubsection.}{1em}{}
\titleformat{\subsubsection}{\normalfont\normalsize\bf\raggedright}{\thesubsubsection.}{1em}{}
\renewcommand\thesection{\arabic{section}}
\renewcommand\thesubsection{\thesection.\arabic{subsection}}
\renewcommand\thesubsubsection{\thesubsection.\arabic{subsubsection}}
\usepackage{epstopdf}
\usepackage[utf8]{inputenc}

\usepackage{hyperref}
\usepackage{xstring}

\usepackage{color}

\title{Generating Major Types of Chinese Classical Poetry \\in a Uniformed Framework}

\name{Jinyi Hu, Maosong Sun$^{*}$ \thanks{$*$\quad Corresponding author}}

\address{ \\Department of Computer Science and Technology, Tsinghua University, Beijing, China\\
Institute for Artificial Intelligence, Tsinghua University, Beijing, China\\
State Key Lab on Intelligent Technology and Systems, Tsinghua University, Beijing, China\\
         hujy17@mails.tsinghua.edu.cn, sms@tsinghua.edu.cn}

\abstract{
Poetry generation is an interesting research topic in the field of text generation. As one of the most valuable literary and cultural heritages of China, Chinese classical poetry is very familiar and loved by Chinese people from generation to generation. It has many particular characteristics in its language structure, ranging from form, sound to meaning, thus is regarded as an ideal testing task for text generation. In this paper, we propose a GPT-2 based uniformed framework for generating major types of Chinese classical poems. We define a unified format for formulating all types of training samples by integrating detailed form information, then present a simple form-stressed weighting method in GPT-2 to strengthen the control to the form of the generated poems, with special emphasis on those forms with longer body length. Preliminary experimental results show this enhanced model can generate Chinese classical poems of major types with high quality in both form and content, validating the effectiveness of the proposed strategy. The model has been incorporated into Jiuge, the most influential Chinese classical poetry generation system developed by Tsinghua University \cite{zhipeng2019jiuge}.
}

\begin{document}

\maketitleabstract

\section{Introduction}

Chinese poetry is a rich treasure in Chinese traditional culture. For thousands of years, poetry is always considered as the crystallization of human wisdom and erudition by Chinese people and deeply influences the Chinese history from the mental and cultural perspective.

In general, a Chinese classical poem is a perfect combination of three aspects, i.e., form, sound, and meaning. Firstly, it must strictly obey a particular form which specifies the number of lines (i.e., sentences) in the poem and the number of characters in each line. Secondly, it must strictly obey a particular sound pattern which specifies the sound requirement for each character in every position of the poem. Lastly, it must be meaningful, i.e., with grammatical and semantic well-formedness for each line and, with thematic coherence and integrity throughout the poem. These three points form the universal principles for human poets to create Chinese classical poems. 

Chinese Classical poetry can be classified into two primary categories, SHI and CI. According to the statistical data from CCPC1.0, a Chinese Classical Poetry Corpus consisting of 834,902 poems in total (We believe it is almost a full collection of Chinese Classical poems). 92.87\% poems in CCPC1.0 fall into the category of SHI and 7.13\% fall into the category of CI. SHI and CI can be further divided into many different types in terms of their forms. We briefly introduce the related background knowledge as follows.

\subsection{SHI}
The majority of SHI has a fixed number of lines and a fixed and identical number of characters for all lines. Two major forms of SHI are \textit{Jueju} and \textit{Lvshi} with four lines and eight lines accordingly. \textit{Jueju} and \textit{Lvshi} are further divided into \textit{Wuyan} \textit{Jueju} and \textit{Qiyan} \textit{Jueju} as well as \textit{Wuyan} \textit{Lvshi} and \textit{Qiyan} \textit{Lvshi} where \textit{Wuyan} means five characters each line and \textit{Qiyan} means seven characters. Figure 1 is a famous classical poem of \textit{Wuyan} \textit{Jueju}. In addition, \textit{Lvshi} has a strict requirement for the two-sentence pairs composed of $<$the third line, the fourth line$>$ and $<$the fifth line, the sixth line$>$: they must satisfy the requirement of \textit{Duizhang}, this is, a strict parallel matching for both part of speech and sense of every character in two lines. This obviously increases the difficulty of poem composition. 

According to CCPC1.0, \textit{Wuyan} \textit{Jueju}, \textit{Qiyan} \textit{Jueju}, \textit{Wuyan} \textit{Lvshi}, and \textit{Qiyan} \textit{Lvshi} constitute 67.96\% of SHI, with 4.26\%, 22.57\%, 15.99\%, and 25.14\% respectively.

\begin{figure}[t]
\centering
\includegraphics[scale=0.47]{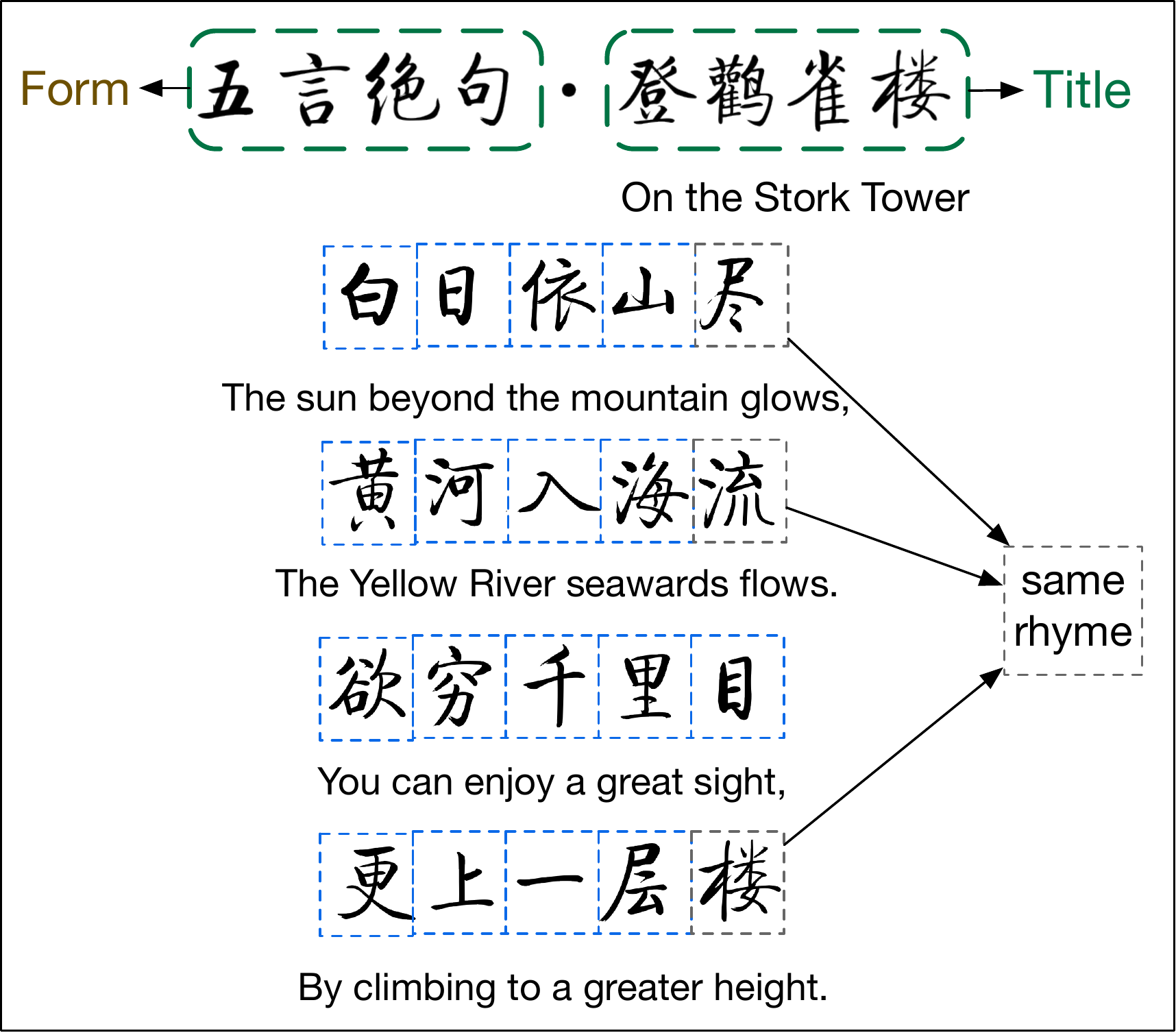}
\caption{An example of SHI with \textit{Wuyan} \textit{Jueju} as its form. The array of small boxes, usually each surrounds a Chinese character, illustrates the form requirement in the number of lines and the number of characters per line for a poem. A full correspondence between character and box, no more and no less, indicates this basic form requirement is satisfied by the given poem.}
\end{figure}

\subsection{CI}
CI is another primary type of Chinese poetry. In contrast to SHI, CI has nearly one thousand forms. Each form of CI (it is called Cipai scholarly) is defined by a fixed number of lines for the poem and, a fixed number of characters for a particular line which usually varies for different lines. The above settings for different Cipai are very distinct, for instance, the Cipai of \textit{Busuanzi} contains 8 lines and 44 characters, as shown in Figure 2, whereas the Cipai of \textit{Manjianghong} contains 22 lines and 94 characters. The high diversity regarding the forms of CI further significantly increases the difficulty of poem composition. 

We observe the statistical distribution of all the forms (Cipai) of CI over CCPC1.0. It roughly follows Zipf’s law \cite{zipf1949human}. There exists a long tail in the distribution where a lot of Cipai only has a few instances which are far less enough for a computational model (algorithm) to learn its forms. So we choose the top frequent 121 forms of CI, constituting 80\% of CCPC1.0, as the focus for CI in this research. 

As can be seen from the above analysis, the greatest challenge for machine generation of Chinese classical poems lies in how to make machine capable of following the universal principles underlying the writing of Chinese classical poems. The to-date research cannot deal with this challenge well. Most of the work so far mainly targeted at automatic generation of \textit{Jueju} (including \textit{Wuyan} \textit{Jueju} and \textit{Qiyan} \textit{Jueju}), for an obvious reason that it is much easier for an algorithm to handle the requirements of form, thematic coherence and integrity in the scenario of four lines than that in the scenario of \textit{Lvshi} with eight lines, let alone much more complicated scenarios, i.e., CI, are taken into account. In fact, the research on the automatic generation of CI is just at the very beginning stage.

In this paper, we propose a uniformed computational framework that tries to generate major types of Chinese classical poems with two major forms of SHI, \textit{Jueju}, and \textit{Lvshi}, as well as 121 major forms (Cipai) of CI using a single model. Preliminary experimental results validate the effectiveness of the proposed framework. The implemented model has been incorporated into Jiuge \cite{zhipeng2019jiuge}, the most influential Chinese classical poetry generation system developed by Tsinghua University (refer to \url{http://jiuge.thunlp.cn/}).
\begin{figure}[!t]
\centering
\includegraphics[scale=0.38]{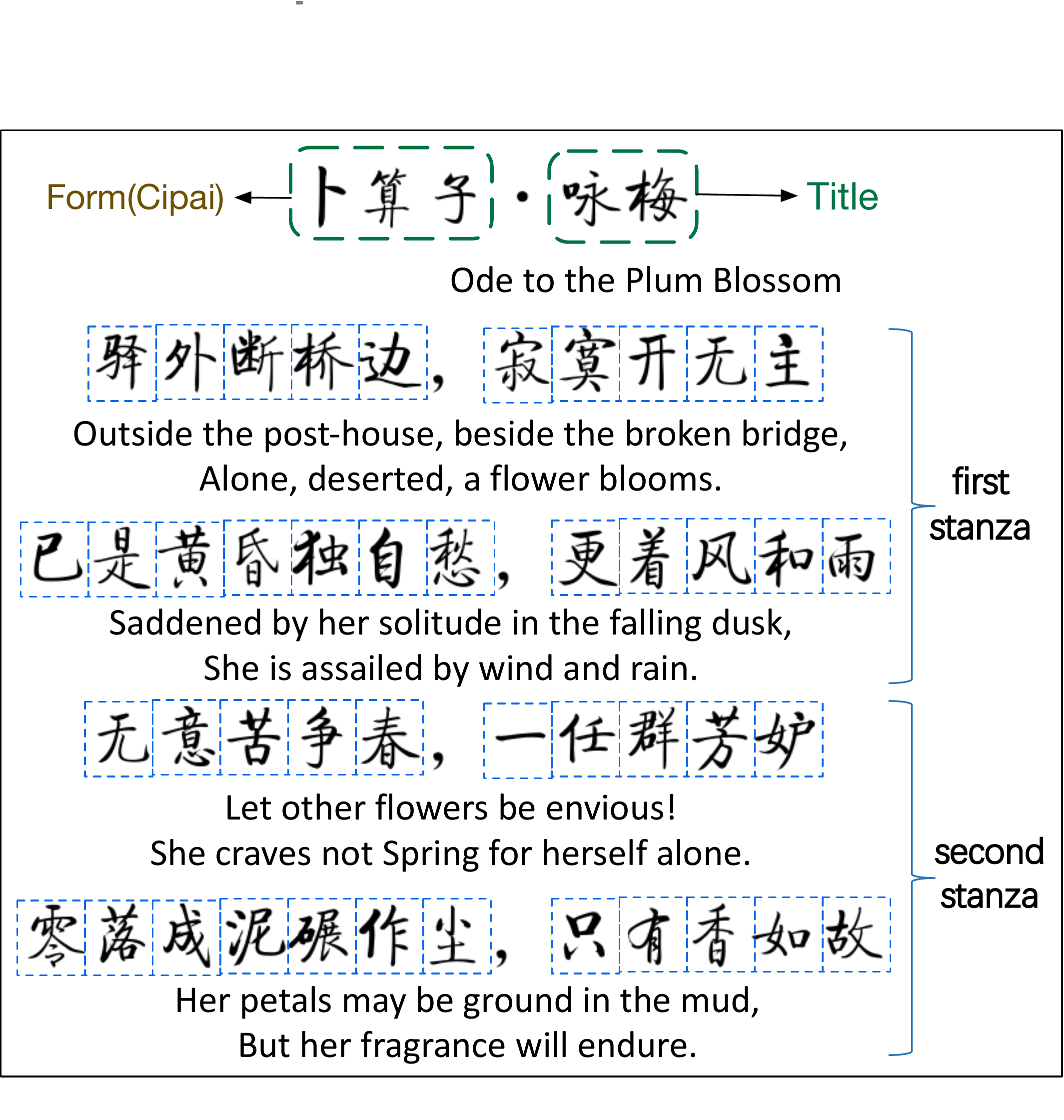}
\caption{An example of CI with the form(Cipai) \textit{Busuanzi}. In contrast to the case of SHI in Figure 1, the array of small boxes here shows the predefined number of characters per line of CI tends to be variable.}

\end{figure}
\section{Related Work}
With the development of deep learning, the mainstream of poem generation research has been shifted from traditional statistical models to neural network methods in recent years. Most existing works are based on the Encoder-Decoder architecture \cite{sutskever2014sequence}. In Chinese classical poetry generation, \newcite{yan2013poet} proposed a model using the Encoder-Decoder architecture and \newcite{wang2016chinese} further used attention-based sequence-to-sequence model.

The key factor in designing the model architecture is how to treat the generated context so far in the process of generating a poem. The input to the encoder could be as short as a single poetic line or all the previously generated lines (whole history). Theoretically, considering the whole history is more appropriate for keeping the thematic coherence and integrity of the generated poem than considering the short history, at the expense that may hurt the fluency of the generated sentences due to the data sparseness problem possibly caused by the more sophisticated model.

Thus we have two basic ways to figure out the history. One is to consider the whole history. \newcite{zhang2014chinese} first introduced the neural network method into poetry generation by proposing the so-called incremental Recurrent Neural Network, where every sentence (line) is embedded into a sentence vector by a Convolutional Sentence Model and then all are packed into a history vector. \newcite{yi2018chinesea} presented a working memory mechanism in LSTM, designing three kinds of memory to address the whole history. Another is to select part of history. \newcite{yi2018chineseb} observed that considering the full context may not lead to good performance in LSTM, and proposed salient clue mechanism where only salient characters in partial history are under consideration.

\begin{figure}[t]
\centering
\includegraphics[scale=0.48]{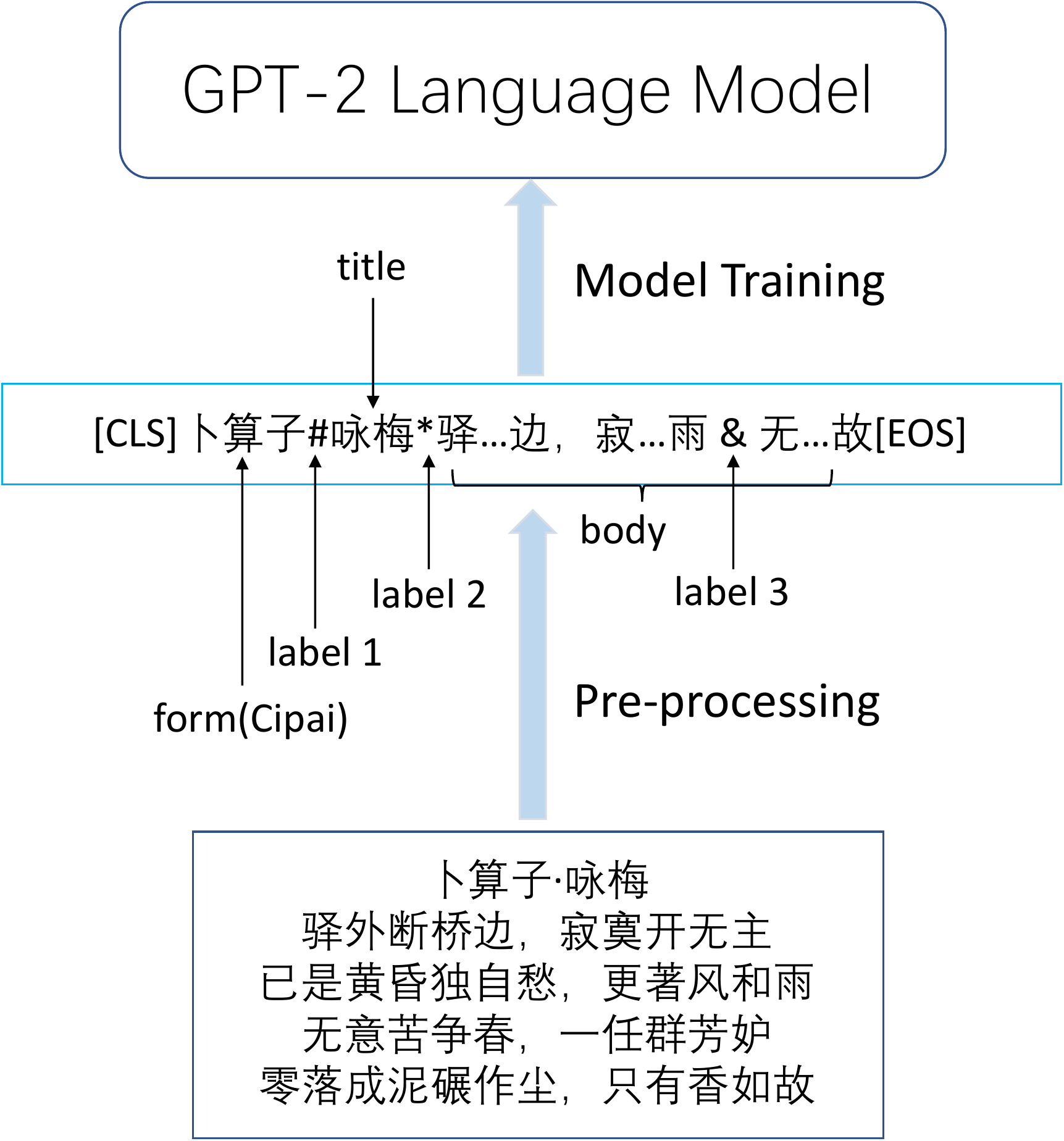}
 
\caption{Format pre-processing of poem samples for training.}

\end{figure}
The Transformer \cite{vaswani2017attention} architecture and other models based on this, including GPT \cite{radford2018improving}, Bert \cite{devlin2019bert}, show much better results in various NLP tasks. Transformer utilizes the self-attention mechanism in which any pair of tokens in the sequence can attend to each other, making it possible to generate much longer SHI or CI while keeping the coherence throughout the poem.

\newcite{liao2019gpt} applied GPT to Chinese classical poetry generation. They pre-trained the model on a Chinese news corpus with 235M sentences and then fine-tuning the model on Chinese poem corpus with 250,000 \textit{Jueju} and \textit{Lvshi}, 20,000 CIs, 700,000 pairs of couplets. A key point is they defined a unified format to formulate different types of training samples, as [\textit{form}, \textit{identifier 1}, \textit{theme}, \textit{identifier 2}, \textit{body}], where “\textit{body}” accommodates the full content of an SHI, CI, or couplet in corresponding “\textit{form}” with “\textit{theme}” as its title. Experiments demonstrated GPT-based poem generation gained promising performance, meanwhile still faced some limitations, for instance, only 70\% of the generated CIs for the Cipai \textit{Shuidiaogetou}, a sort of CI with quite long body, are correct in form.

Regarding this, we think the work of \newcite{liao2019gpt} could be improved in the following three respects. First, there is a large improving room for better fitting the form requirement of CI in the process of generation, especially for those with relatively long body length. Second, their formulation format for training samples can be supplemented, for example, the stanza structure of CI is missing. Third, using contemporary Chinese news corpus to pre-train the model may not be necessary, owing to distinctive differences in both meaning and form between contemporary Chinese and Chinese classical poetry language.
 
For the above considerations, we give up the pre-training on the news corpus and add a separation label to indicate the stanza structure of CI. Then we make use of GPT-2 to train the model. Furthermore, we propose a form-stressed weighting method in GPT-2 to strengthen the control in particular to the form of CI.

\section{Model}
\subsection{Pre-processing}
We present a unified format for formulating all types of training samples of SHI and CI by extending the format given in \newcite{liao2019gpt}. First, we change various punctuations between lines into the comma ‘,’, serving as a uniform separation label between two lines. Second, we utilize three separation labels, $[label_1]$ and $[label_2]$ to separate between form, title, and body of the poem respectively, and $[label_3]$ to separate two stanzas of CI if needed. Third, we enclose $[EOS]$ at the end of the body. Thus, the format for SHI is as follows:
\begin{align*}
    &[CLS]form[label_1]title[label_2]body[EOS] \\
    &body : line_1, line_2,...,line_n
\end{align*}
where n is the number of lines in the poem.

The format of CI will be enriched with $[label_3]$ if it has two stanzas in the body:
\begin{align*}
    &[CLS]form[label_1]title[label_2]body[EOS] \\
    &body : stanza_1[label_3]stanza_2 \\
    &stanza_1 : line_1, line_2,...,line_m  \\
    &stanza_2 : line_{m+1}, line_{m+2},...,line_n 
\end{align*}
Here, $[label_1]$, $[label_2]$ and $[label_3]$ are set as ‘$\#$’, ‘$*$’ and ‘$\&$’.

After pre-processing, all the formatted poem samples will be sent to the poetry generation model for training, as illustrated in Figure 3.   

\begin{figure*}[!t]
\begin{subfigure}{0.45\textwidth}
\includegraphics[width=\linewidth]{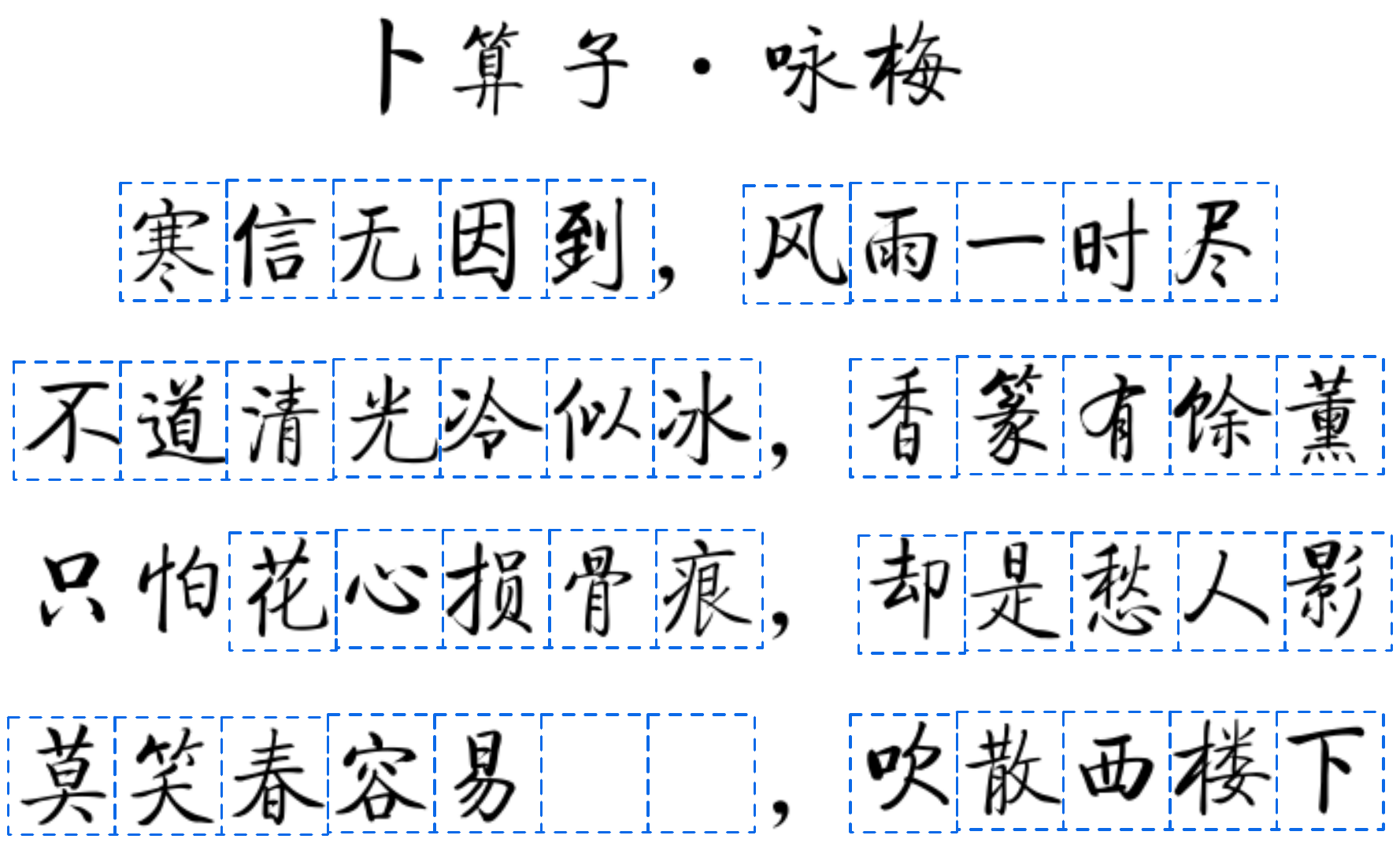}
\caption{A generated poem by the basic model: two obvious errors in form.} \label{fig:1a}
\end{subfigure}
\hspace*{\fill} % separation between the subfigures
\begin{subfigure}{0.45\textwidth}
\includegraphics[width=\linewidth]{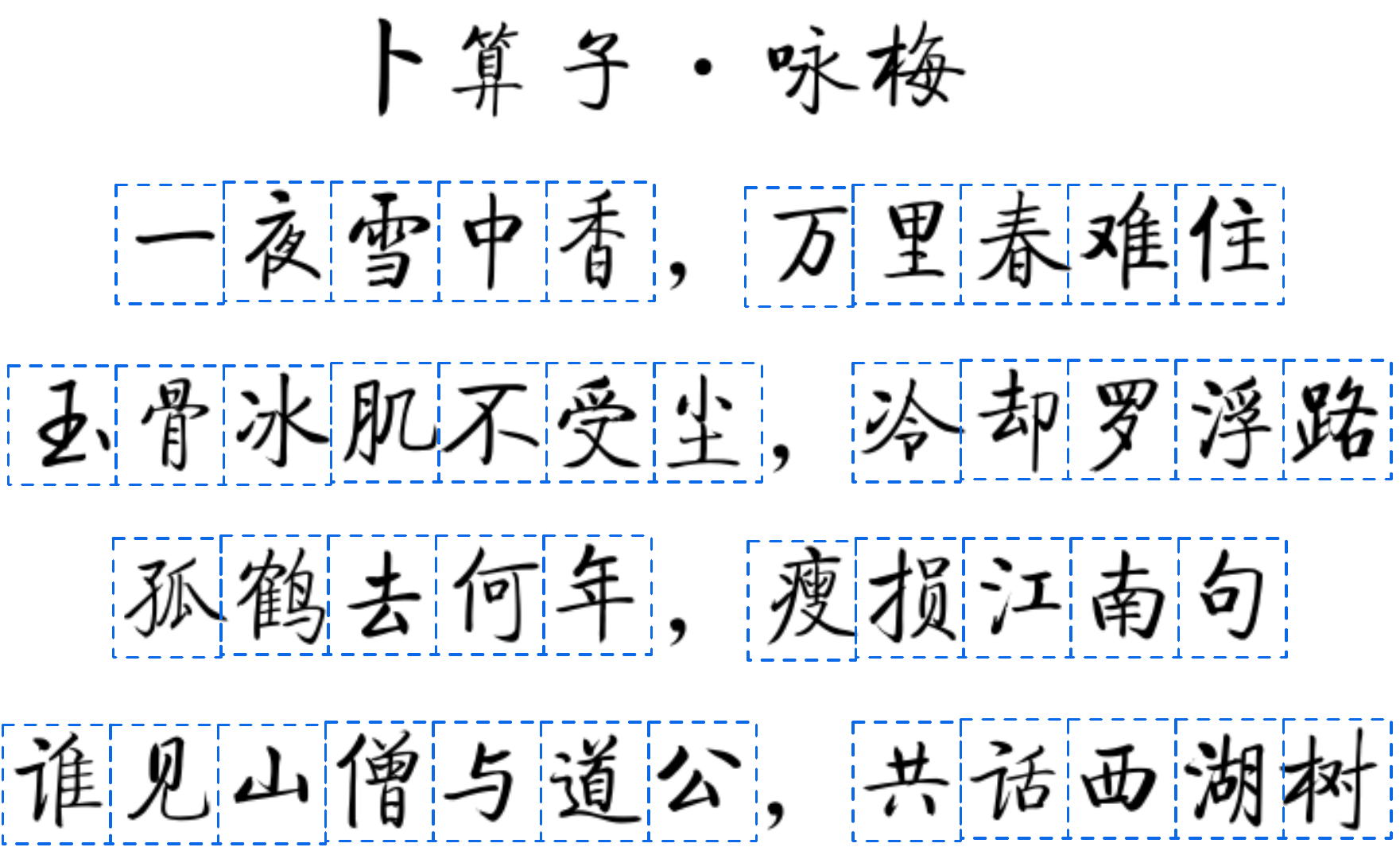}
\caption{A generated poem by the enhanced model, with the same inputting title (or theme) under the same Cipai as in (a): full correctness in form.e} \label{fig:1b}
\end{subfigure}
\caption{Comparison of two generated poems by the basic model and the enhanced model.} \label{fig:1}
\end{figure*}

\subsection{Basic Model}
We leverage the Transformer-based GPT-2, which is often used to train a robust language model, as the basic model of poetry generation. Compared to previous neural network-based language models such as RNN and LSTM, it is reported that GPT-2 exhibits good performance in the quality of generated texts given quite a long history \cite{radford2019language}. To weaken the so-called degeneration problem in generation and increase the diversity of generated texts, we use the top-k stochastic sampling strategy \cite{fan2018hierarchical} (k is set as 15 in our experiment) to choose the next tokens to generate. In addition, our poetry generation model takes the Chinese character rather than the word as a basic linguistic unit, so word segmentation is not needed.

With this naive GPT-2 model, we see from the experimental results that the generated poems appear pretty good in both meaning and sound(including rhyme), though if being observed carefully, there still exist some in-depth problems in sentence fluency and thematic coherence of the whole poem which are uneasy to solve. As for form, the model can perform well in generating \textit{Jueju} and \textit{Lvshi} of SHI whereas rather poorly in generating various Cipai of CI, with quite high form errors. Figure 4(a) is an example of a generated CI by this model, under Cipai of \textit{Busuanzi}, where two characters are mistakenly missing which obviously violates the form requirement.

\subsection{Enhanced Model}
In the basic model, the loss function for training with respect to the $i$th token in the text is conventionally defined as the cross-entropy:
\begin{align*}
    Loss(x,i) &= -\log\frac{\exp{x[i]}}{\sum\limits_{j}\exp{x[j]}} \\
        &= - x[i] + \log\sum\limits_{j}\exp{(x[j])} 
\end{align*}
where $x[i]$ is the vector of $i$th token, $j$ is over all possible token types.

To address the form problem, we simply add a weighting factor into the loss function with particular stress on the aforementioned three types of form-related tokens, i.e., the line separation label ‘,’, the stanza separation label ‘$\&$’, and $[EOF]$, as in:

\begin{align*}
    Loss(x,i) = weight[i] \big(- x[i] + \log\sum\limits_{j}\exp{(x[j])}  \big )
\end{align*}
where $weight[i]$ is set as 1 for any Chinese character, 2 for ‘,’ and ‘$\&$’, and 3 for $[EOF]$.

This simple method (we thus call it the form-stressed weighting method) enhances the model’s capability to form control quite significantly. Figure 4(b) shows an example that contrasts the case in Figure 4(a).

\section{Experiment}
\begin{table*}[h]
\renewcommand\arraystretch{1.7}
\centering

\begin{tabular}{|c||c||c||c||c|}
\hline
Cipai & Length of Body & \makecell[c]{Number of \\Training Samples} & \makecell[c]{Correct Rate in Form of \\
Basic model} & \makecell[c]{Correct Rate in Form of \\
Enhanced model} \\
\hline
\textit{Rumengling} & 33 &682 &  86.0\% & 90.0\% \\
\hline
\textit{Jianzimulanhua} &44 &866 &  87.3\% & 95.7\% \\
\hline
\textit{Qingpingyue} & 46 & 1236 & 84.0\% & 96.0\%  \\
\hline
\textit{Dielianhua} &60 & 1578 &  89.7\% & 91.3\%  \\
\hline
\textit{Manjianghong} & 93 &1398 &  42.1\% & 83.3\% \\
\hline 
\textit{Qinyuanchun} & 114 &1061 &  12.0\% & 55.0\% \\
\hline
\end{tabular}
\caption{Comparison between two models on the control to the form of CI.}
\end{table*}

\begin{figure*}[!t]
\begin{subfigure}{0.48\textwidth}
\includegraphics[width=\linewidth]{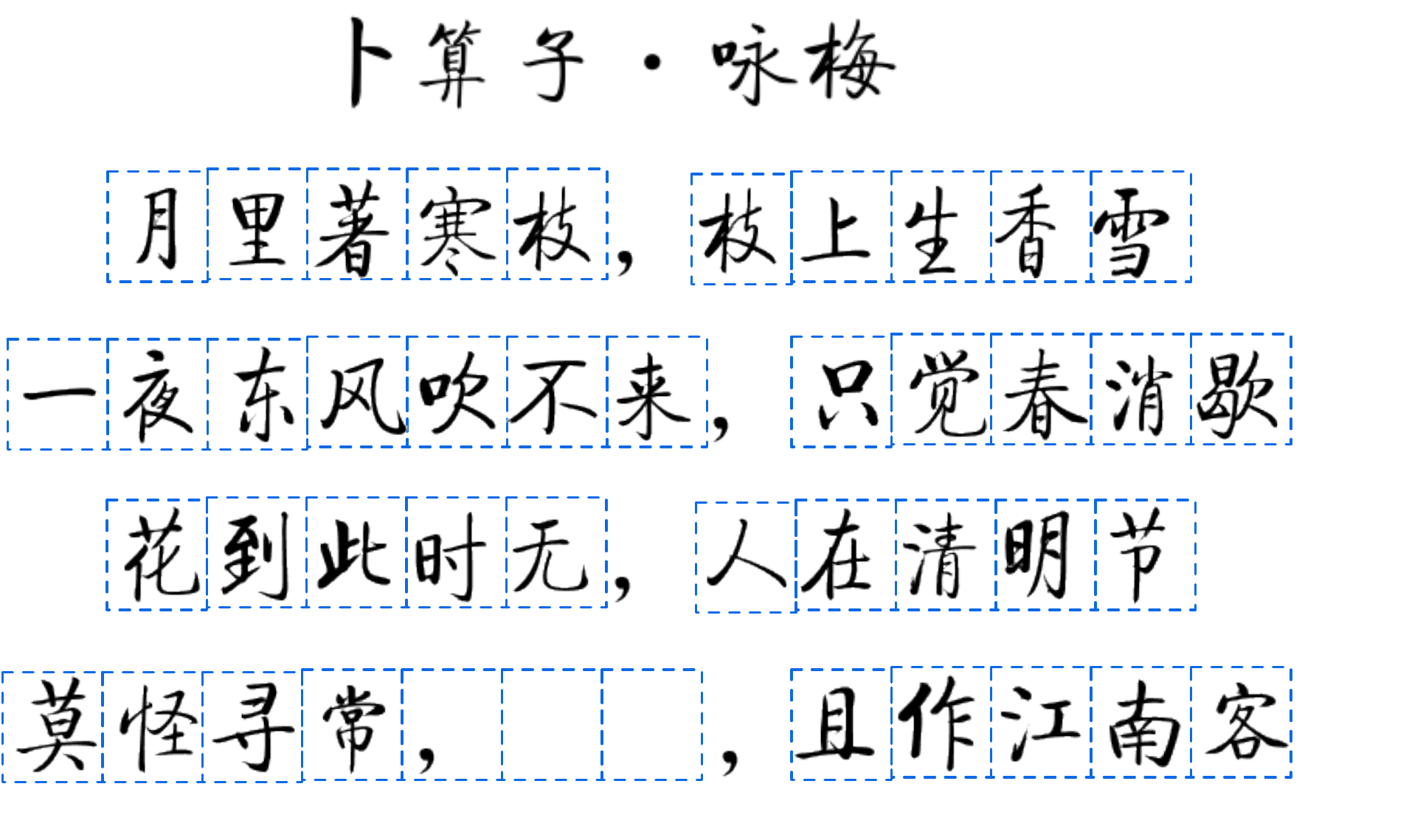}
\end{subfigure}
\hspace*{\fill} % separation between the subfigures
\begin{subfigure}{0.48\textwidth}
\includegraphics[width=\linewidth]{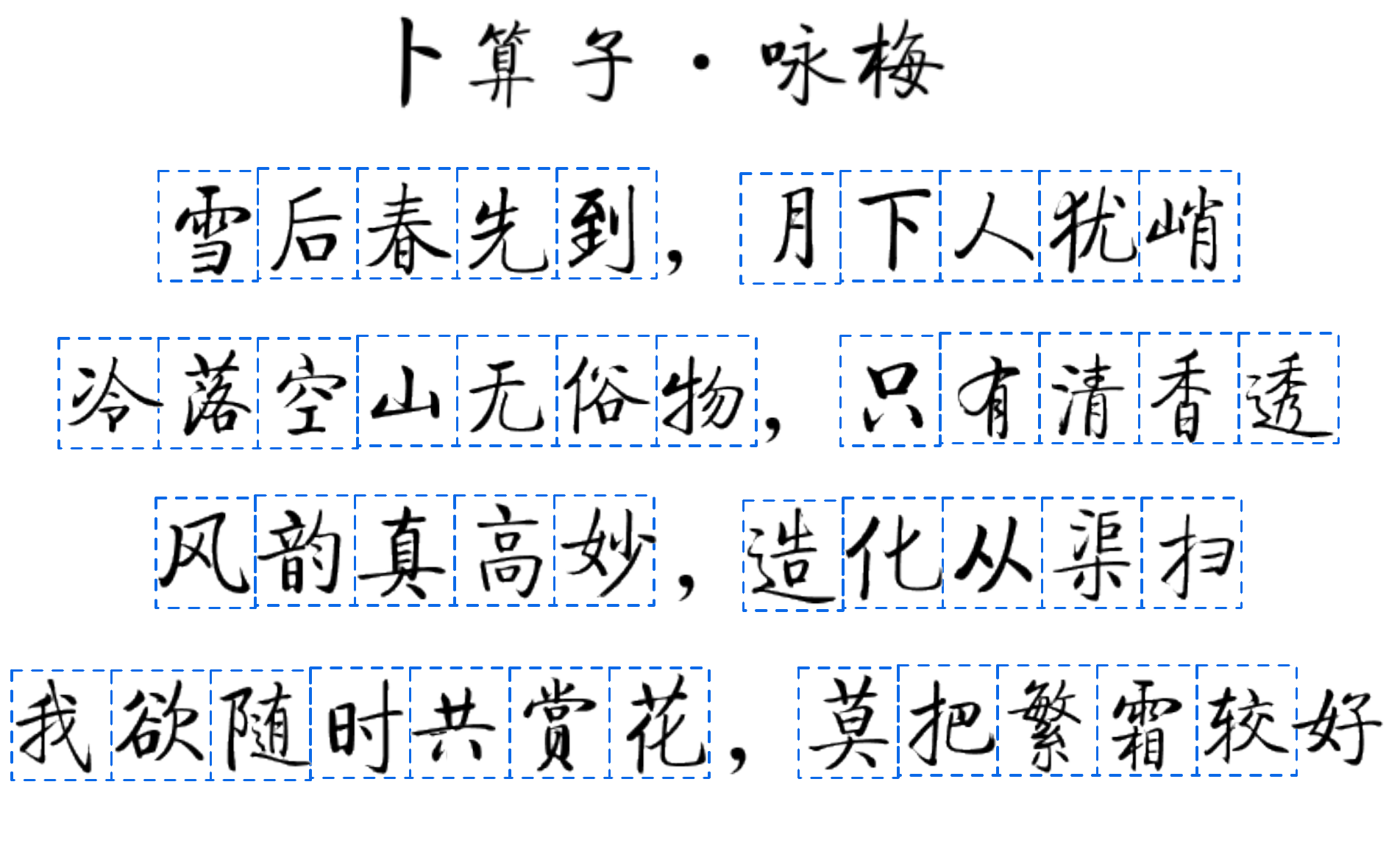}
\end{subfigure}
\caption{Two example poems generated by the model without considering the stanza separation. Both have errors in form. Refer to Figure 4(b) for comparison.} \label{fig:1}
\end{figure*}

\subsection{Experiment Setup}
We implement the GPT-2 model based on the transformers library \cite{Wolf2019HuggingFacesTS}. The model configuration is 8 attention heads per layer, 8 layers, 512 embedding dimensions, and 1024 feed-forward layer dimensions. We employ the OpenAIAdam optimizer and train the model with 400,000 steps in total on 4 NVIDIA 1080Ti GPUs. The characters with frequency less than 3 in CCPC1.0 are treated as UNK and a vocabulary with 11259 tokens (characters) is finally built up.

\subsection{Performance Comparison of the Two Models in Form}
For \textit{Jueju} and \textit{Lvshi} of SHI, because of their simplicity in form, the two models hardly make form errors. We generate 500 poems for each type using the two models accordingly. All of these poems are in the right form. This demonstrates that both models are all very powerful in generating \textit{Jueju} and \textit{Lvshi} with almost perfect performance in form.

For CI, we select 6 Cipais, with the body length varying from 33 to 114 characters and with relatively sufficient training samples in CPCC, as our observation target. We generate 300 poems with the two models accordingly. Table 1 summarizes the correct rates of the two models under these 6 Cipais (a generated poem is considered to be correct in form if and only if its form fully matches the expected form). As can be seen, a tendency is the longer the body of CI, the worse the performance of the two models in form and, the more significant the gain in the form correct rate for the enhanced model (an extreme is in the case of \textit{Qinyuanchun} where the correct rate is raised from 12.0\% to 55.0\%).

\subsection{Effect of the Stanza Separation}
The preliminary observation on the generated poems suggests that the inclusion of the stanza separation into the unified format of training samples is beneficial in some degree for meeting the form requirement. For instance, we input the same title to the enhanced model and to a model trained under the same condition except without the stanza separation, asking them to generate a number of CIs with Cipai of \textit{Busuanzi}, a task similar to that in Figure 4. We find that about 20\% of CIs generated by the latter suffer from some errors in form, as illustrated in Figure 5, meanwhile all the CIs generated by the former ideally match the expected form.

\subsection{Case Observation}
According to our observation, the enhanced model is likely to generate poems with both high quality and diversity. We present two examples generated by the model and give some comments on the meaning of each poem.

\begin{center}
    \begin{CJK*}{UTF8}{gbsn}七律 · 远望\end{CJK*} \\
    \begin{CJK*}{UTF8}{gbsn}江上微茫一叶舟，天涯芳草满汀洲\end{CJK*} \\
    \begin{CJK*}{UTF8}{gbsn}数声渔唱隔船过，几点人家落帆游\end{CJK*} \\
    \begin{CJK*}{UTF8}{gbsn}春色不从莺语到，夕阳空度客心愁\end{CJK*} \\
    \begin{CJK*}{UTF8}{gbsn}何时重向长桥饮，同泛溪光共白头\end{CJK*} 
\end{center}

The example above is a \textit{Qiyan} \textit{Lvshi}. The title of this poem means “look far around”. In this poem, the first four lines depict a view seen from the river bank-misty and rolling waters, a drifting boat, lush vanillas, melodies from passing boats and cottages on the bank, creating a tranquil and halcyon atmosphere. However, the poet is still overcome by solitude and nostalgia because of the lonely trip, which is vividly revealed in the second four sentences. The poem adopts a typical semantic structure of \textit{Qiyan} \textit{Lvshi} with its first-half delineating a view and then conveying the poet’s feeling in the second-half (the contrast between the view and the feeling is one of the appreciated artistic methods in Chinese classical poems). In addition, for \textit{Lvshi}, the pairs of $<$the third line, the fourth line$>$ and $<$the fifth line, the sixth line$>$ must satisfy the requirement of \textit{Duizhang}, a correspondence in both part-of-speech(POS) and word sense between two parallel lines. This point is perfectly reflected in the generated poem, as shown in Table 2.

\begin{table*}[h]
\centering
\renewcommand\arraystretch{1.5}

\begin{tabular}{c|cccccc}
\hline
        & \begin{CJK*}{UTF8}{gbsn}数\end{CJK*}  & \begin{CJK*}{UTF8}{gbsn}声\end{CJK*}  & \begin{CJK*}{UTF8}{gbsn}渔歌\end{CJK*}  & \begin{CJK*}{UTF8}{gbsn}隔\end{CJK*}  & \begin{CJK*}{UTF8}{gbsn}船\end{CJK*}  &  \begin{CJK*}{UTF8}{gbsn}过\end{CJK*} \\
POS     & NUMBERAL  & N  & N  & V  & N  & V  \\
Word    &  several & sound & fishing song & next-door & boat & pass\\
Meaning & \multicolumn{6}{c}{\makecell[c]{The next-door boat is passing by, \\with several sounds of a fishing song}} \\
\hline
        & \begin{CJK*}{UTF8}{gbsn}几\end{CJK*}  & \begin{CJK*}{UTF8}{gbsn}点\end{CJK*}  & \begin{CJK*}{UTF8}{gbsn}人家\end{CJK*}  & \begin{CJK*}{UTF8}{gbsn}落\end{CJK*}  & \begin{CJK*}{UTF8}{gbsn}帆\end{CJK*}  &  \begin{CJK*}{UTF8}{gbsn}游\end{CJK*} \\
POS     & NUMERAL &   N   & N  & V  & N  & V  \\
Word    & a few & point &  home & fall & sail & move \\
Meaning & \multicolumn{6}{c}{\makecell[c]{A few of far-away boats which carry on the whole family, \\ still wander with the falling sails, looking like small pieces of points}} \\
\hline
\end{tabular}
\caption{Illustration of \textit{Duizhang}.}
\end{table*}

\begin{center}
    \begin{CJK*}{UTF8}{gbsn}满江红 · 塞外\end{CJK*} \\
    \begin{CJK*}{UTF8}{gbsn}风急秋空，天欲暮，黄云飞处。\end{CJK*} \\
    \begin{CJK*}{UTF8}{gbsn}人不见，沙堤野戍，乱鸦啼苦。\end{CJK*} \\
    \begin{CJK*}{UTF8}{gbsn}万里胡笳吹雁断，三更羌笛愁如许。\end{CJK*} \\
    \begin{CJK*}{UTF8}{gbsn}甚关河、征妇泪痕多，无行路。\end{CJK*} \\
    \begin{CJK*}{UTF8}{gbsn}青狼火，荒烟树。\end{CJK*} \\
    \begin{CJK*}{UTF8}{gbsn}白露草，残阳度。\end{CJK*} \\
    \begin{CJK*}{UTF8}{gbsn}但寒山远近，故乡千古。\end{CJK*} \\
    \begin{CJK*}{UTF8}{gbsn}一角斜晖归梦绕，满江红叶西陵去。\end{CJK*} \\
    \begin{CJK*}{UTF8}{gbsn}待明年，又到汉家城，重回顾。\end{CJK*} \\
\end{center}

The example above is a CI in the form of \textit{Manjianghong} and the title means “beyond the Great Wall”. It vividly depicts a typical view of the Northwestern China howling wind, clouds of dust, crying crows and lugubrious sound of flutes. The poem is saturated with nostalgia, solitude and desolate feelings of life, which is not only embodied in the bleak scenery but also overtly revealed in the last three sentences. The combination of visual and audio feelings and of reality and imagination is tactfully employed in the poem and makes it even more impressive and resonating.

\section{Conclusion and Future Works}
In this paper, we propose a GPT-2 based uniformed framework for generating major types of Chinese classical poems, including SHI and CI. To this end, we at first define a unified format for formulating all types of training samples by integrating more detailed form information, then present a simple form-stressed weighting method in GPT-2 to strengthen the control to the form of CI. Preliminary experiments validate the effectiveness of our method. Nevertheless, we also find that enabling GPT-2 to have a strong capability in form manipulation for the generated texts remains a difficult challenge, particularly for those forms with longer body length and fewer training samples. We plan to figure out a more sophisticated way to make the model better learn the form structure and hope to enrich the general GPT-2 from this special perspective.

\section{Acknowledgements}
We would like to thank Zhipeng Guo, Xiaoyuan Yi, Xinran Gu and anonymous reviewers for their insightful comments. This work is supported by the project Text Analysis and Studies on Chinese Classical Literary Canons with Big Data Technology under grant number 18ZDA238 from the Major Program of the National Social Science Fund of China. Hu is also supported by the Initiative Scientific Research Program and Academic Training Program of the Department
of Computer Science and Technology, Tsinghua University.

\section{References}\label{reference}
%\label{main:ref}
%\bibliographystyle{unsrt}
\bibliographystyle{lrec}
\bibliography{lrec2020W-xample-kc}
\end{document}